\documentclass{article}
\usepackage{spconf,amsmath,epsfig}
\usepackage{multirow}
\usepackage{amssymb}

\usepackage{bm}
\usepackage{multicol}
\usepackage{amsfonts}
\usepackage{paralist}
\usepackage{amsthm}
\usepackage{indentfirst}
\usepackage{mathrsfs}
\usepackage{indentfirst}
\usepackage{graphicx}
\usepackage{booktabs}
\usepackage{subfigure}
\usepackage{lipsum}
\usepackage{graphics}

\usepackage[switch,pagewise,columnwise]{lineno}
\usepackage[noadjust]{cite}

\let\OLDthebibliography\thebibliography
\renewcommand\thebibliography[1]{
  \OLDthebibliography{#1}
  \setlength{\parskip}{0pt}
  \setlength{\itemsep}{0pt plus 0.3ex}
}

\begin{document}\sloppy

% Example definitions.
% --------------------
\def\x{{\mathbf x}}
\def\L{{\cal L}}

% Title.
% ------
\title{Series Photo Selection via Multi-view Graph Learning}
%
% Single address.
% ---------------

% \name{Jin Huang\IEEEauthorrefmark{1},Lu Zhang\IEEEauthorrefmark{2},Yongshun Gong\IEEEauthorrefmark{1},Jian Zhang\IEEEauthorrefmark{3},Xiushan Nie\IEEEauthorrefmark{4},Yilong Yin\IEEEauthorrefmark{1}}
% \IEEEauthorblockA{\IEEEauthorrefmark{1}Shandong University}
% \IEEEauthorblockA{\IEEEauthorrefmark{2}Twentieth Century Fox, Springfield, USA}
% \IEEEauthorblockA{\IEEEauthorrefmark{3}Starfleet Academy, San Francisco, CA 96678 USA}
% \IEEEauthorblockA{\IEEEauthorrefmark{4}Shandong Jianzhu University}

\name{\begin{tabular}{c}Jin Huang$^{1}$, Lu Zhang$^{2}$, Yongshun Gong$^{1 \star}$, Jian Zhang$^{3}$, Xiushan Nie$^{4}$, Yilong Yin$^{1 \star}$\end{tabular}
\thanks{Corresponding author: Yongshun Gong and Yilong Yin.}
}

\address{$^{1}$ School of Software, Shandong University\\
        $^{2}$ University of Queensland  \\
        $^{3}$ University of Technology Sydney \\
        $^{4}$ School of Computer Science and Technology, Shandong Jianzhu University}
%Address and e-mail should NOT be added in the submission paper. They should be present only in the camera ready paper. 

\maketitle

\begin{abstract}
Series photo selection (SPS) is an important branch of the image aesthetics quality assessment, which focuses on finding the best one from a series of nearly identical photos. While a great progress has been observed, most of the existing SPS approaches concentrate solely on extracting features from the original image, neglecting that multiple views, e.g, saturation level, color histogram and depth of field of the image, will be of benefit to successfully reflecting the subtle aesthetic changes. Taken multi-view into consideration, we leverage a graph neural network to construct the relationships between multi-view features. Besides, multiple views are aggregated with an adaptive-weight self-attention module to verify the significance of each view. Finally, a siamese network is proposed to select the best one from a series of nearly identical photos. Experimental results demonstrate that our model accomplish the highest success rates compared with competitive methods.
\end{abstract}

\section{Introduction}
% \label{sec:intro}

With the rapid development of multimedia technology, especially the miniaturization and portability of photographic equipment, humans are more inclined to use digital devices to record their lives. Coming along with this, the number of photos taken by users has increased rapidly. Sometimes people will press the shutter repeatedly for the worry of missing fleeting moments, resulting in several photos taken in 
a short time interval. Users have to browse dozens or even hundreds of photos to organize them and clear digital memories. %In addition, for the subject of photography is the beautiful building, different users may also take repeated photos, which also poses a very serious challenge to the recommendation and management of image search engines. 
Faced with this phenomenon, automatic series photo selection (SPS) is of great significance.

SPS is a branch task in the field of image aesthetics quality assessment. Some prior aesthetic quality assessment studies use hand-designed photographic features \cite{DattaJLW06, LuoT08}, general descriptors \cite{MarchesottiPLC11, SuCKHC11, PerronninD07}, and deep features \cite{LiuPKB20, HosuGS19, KongSLMF16, ChenZZLXZ020} to solve aesthetic classification or regression problems. These features are not comprehensive and accurate enough for the abstraction of image information. For example, due to the vagueness of certain photography or artistic rules, manual features cannot be approximated; since the input image needs to be reshaped using deep learning models, the original beauty of the image may be damaged. As shown in Figure 1, the image with the highest score predicted by an advanced aesthetic quality assessment model \cite{LuLJYW14} may not be the best image in the series. Even there have been some efforts to overcome the limitations of deep method, existing methods only extract the deep features, while neglecting that multiple views, e.g, saturation level and color histogram, will also be of benefit to reflecting the subtle aesthetic changes. 

\begin{figure}[t]
\centering
\label{1}
\includegraphics[width=0.5 \textwidth]{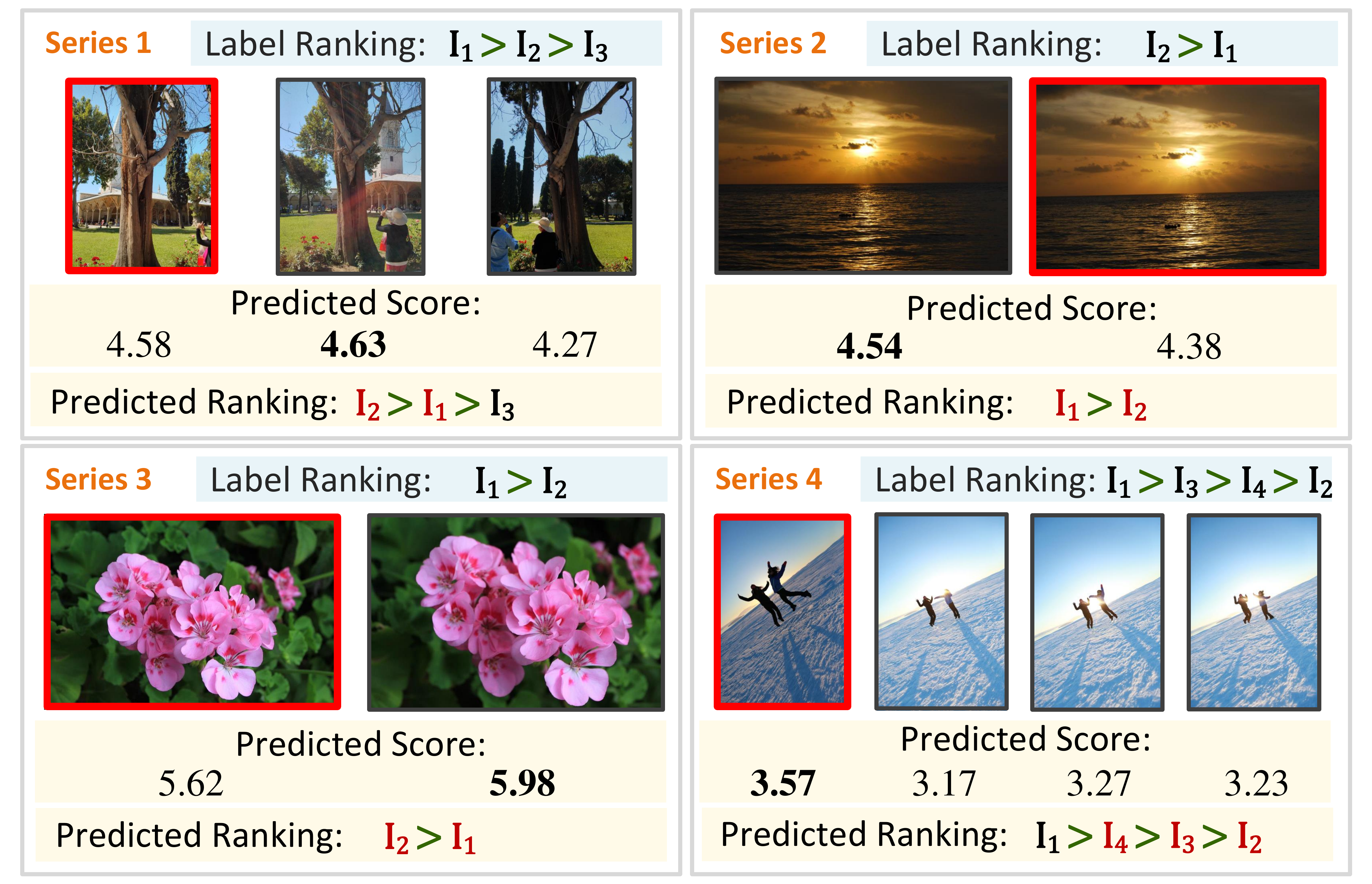}
\caption{Examples of the SPS task. SPS aims to select the photo with the best aesthetic quality from a series set, labeled with the red box. In some cases, existing methods cannot predict the correct label ranking because of the inadequacy of the single view representation.}
\end{figure}

In this study, we focus on addressing the SPS problem with the multi-view graph learning. We embed multiple views by Graph Convolution Networks (GCNs) and jointly extract the aesthetic representation from images. One of the major challenges is how to mine the deep spatial complementary information between different views (manual features, general features, deep features, etc.). Especially when extracting multiple views from the same image, there are huge differences in feature extraction strategies, scales and semantic information. A forcible concatenation will introduce a part of negative information, thus affect the complementary representations of different views. Moreover, it is important to select an effective combination methods to integrate all views.

\begin{figure*}[t]
\centering
\label{2}
\includegraphics[width=1 \textwidth]{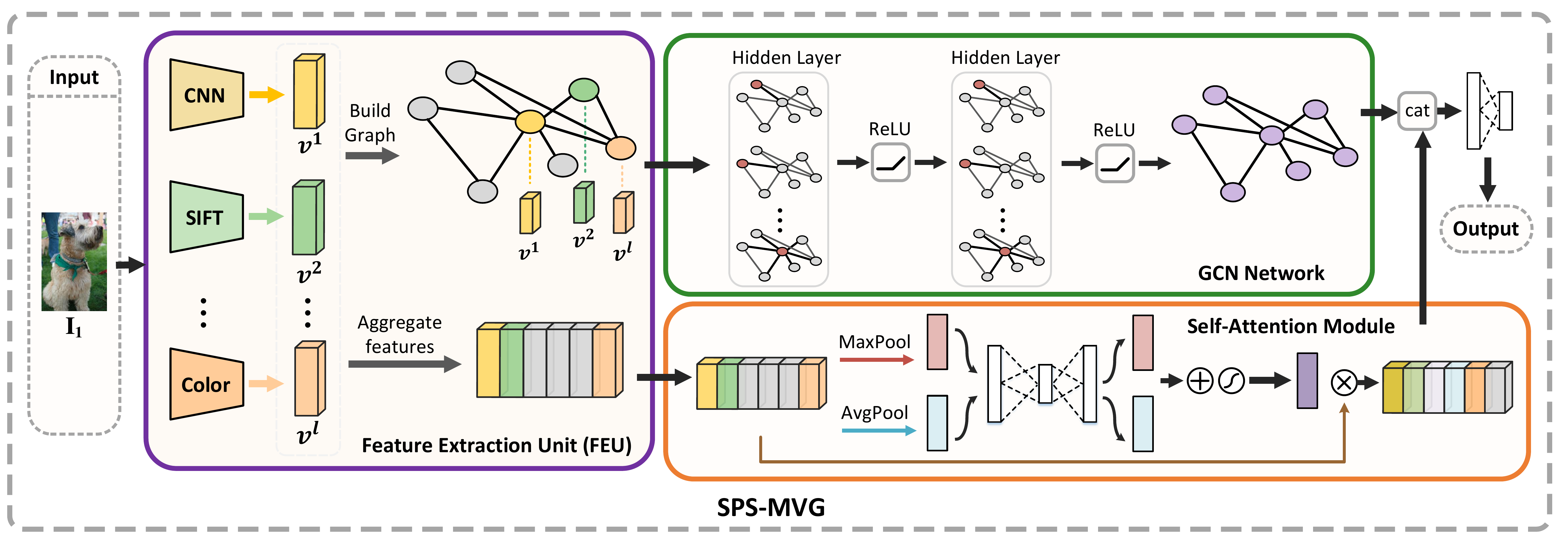}
\caption{The framework of SPS-MVG . It takes one of the image pairs as input, and generates $l$ views through the feature extraction module. Multi-view graph is built based on the similarities. We also utilize these multi-view features to learn the common information through a self-attention module. Finally, two types of features, one with the enhanced spatial information, and the other with the enhanced content are combined as the latent image representation.}
\end{figure*}

To overcome above-mentioned challenges, we propose a novel siamese network for the SPS problem with multi-view graph learning (SPS-MVG). We try to represent an image from different perspectives (multiple views) and fuse those views rationally. To this end, a graphic structure is devised to keep the representation of the image, which not only retains the unique characteristics of the multiple views, but also introduces the spatial correlations between features into the aesthetic assessment process. The motivation behind this is to leverage the consistency and complementarity of various views. The nodes of a graph present different views, and edges are constructed by the similarities between views. Besides, to strengthen the consistent information between views, we also merge different views into a latent representation via an adaptive-weight self-attention module. Finally, we propose a siamese network to assess the aesthetic quality of a pair of images. The main contributions can be summarized as follows:

%Our framework process is as follows: We establish the graphic structure representation of the image, which not only retains the unique characteristics of the feature itself, but also introduces the spatial information between the features into the aesthetic assessment process. Then use the two-layer graph convolutional layer to obtain the embedded features and strengthen the spatial relationship between the features. We also use the self-attention module to further strengthen the common information between features and weaken the feature representations with excessive semantic differences. Finally our framework jointly learns the embedded features of multi-view data containing as much information as possible for series photo selection. We conduct a variety of experiments on a real dataset Phototriage. The results demonstrate our superior performance on this task compared with state-of-the-art methods.

%   (1) To the best of our knowledge, SPS-MVG provides the first attempt on addressing the SPS problem with multi-view graph learning. The graph is designed to express relationships among different feature views.
  
%   (2) We propose an end-to-end multi-view collaborative fusion strategy based on the self-attention module, and propose a siamese network accordingly.
  
%   (3) We conduct extensive experiments on the real-world dataset Phototriage. Experimental results demonstrate the effectiveness of SPS-MVG compared with other competitive models.

\begin{itemize}
  \item To the best of our knowledge, SPS-MVG provides the first attempt on addressing the SPS problem with multi-view graph learning. The graph is designed to express relationships among different feature views.
  \item We propose an end-to-end multi-view collaborative fusion strategy based on the self-attention module, and propose a siamese network accordingly.
  \item We conduct extensive experiments on the real-world dataset Phototriage. Experimental results demonstrate the effectiveness of SPS-MVG compared with other competitive models.
\end{itemize}

\section{Related Work}
%We briefly review the related works in two aspects: series photo selection (SPS) and image aesthetics quality assessment.

To data, very limited studies investigate the series photo selection problem. Chang et al. \cite{Finkelstein16} collected the first large-scale public data set composed of photo series in personal albums, and used the Siamese Network to propose an end-to-end deep learning method. Huang et al. \cite{HuangCZSYY20} aggregated multi-scale features from different network layers and combined features to distinguish series of photos. These SPS works focus only on extracting features from the single view of images.

Early conventional methods of the image aesthetics quality assessment design image layouts by approximating simple photography composition guidelines, e.g., depth of field, rule of thirds, and brightness (exposure) \cite{DattaJLW06, LuoT08, WongL09}. However, the artificially designed features are inspired by photography or psychology, and the scope is limited and it is difficult to guarantee the validity. General image descriptors are used to represent the aesthetic features \cite{MarchesottiPLC11, SuCKHC11, MurrayMP12, PerronninD07} such as SIFT and Fisher Vector, which are used to capture the general features of natural images, rather than specifically used to describe the aesthetics of the image, so they also have great limitations. Generally, images need to be cropped, scaled or filled to meet the requirement of the fixed-size inputs before fed into neural network \cite{MarchesottiPLC11, SuCKHC11, MurrayMP12, PerronninD07}. However, due to the sensitivity of series photo aesthetic evaluation to these indicators, these conversions cannot be applied well in the SPS problem directly.

\section{Proposed Method}
%In this section, we describe the problem series photo selection, and propose our SPS method with multi-view graph learning (SPS-MVG).
The proposed method adopts two-branch networks with shared parameters. The overall framework is shown in Figure 2.
First, we use the feature extraction unit (FEU) to uniformly process the different feature views of the same image $\bm{x}_i$ (including manual features, general features, and deep features, etc.) as multiple views $\bm{V}_i$, then we build the graph structure with $\bm{V}_i$ and concatenate features.
Secondly, we use a two-layer graph convolutional network to automatically learn the feature information and structural information of the graph at the same time. The self-attention module is also used to reinforce the common information of different feature views and maximize the consistency between those views.
Finally, we merge the multi-view features and connect the features of the two images to obtain new representations. After the multilayer perceptron with three full connection layers, a mapping function is obtained, and the cross-entropy loss is used to obtain the comparison results of image pairs.

\begin{figure}[t]
\centering
\label{3}
\includegraphics[width=0.5\textwidth]{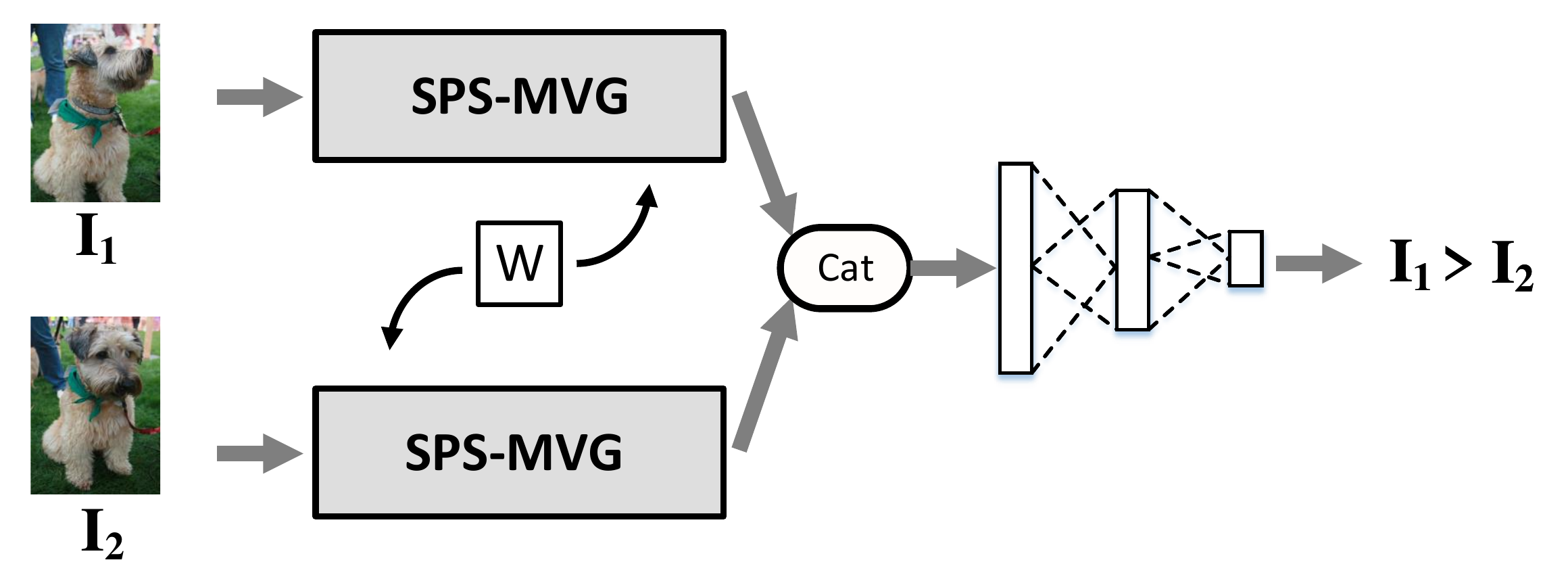}
\caption{Siamese architecture: Features are extracted from each of the pair of
images by two SPS-MVG networks with shared weights. Then the difference of the features are
passed through two levels of hidden layers. Finally a two-way softmax decides which image is preferred.}
\end{figure}

\subsection{Problem Formulation}
We denote $\mathcal{D} = \{ {( {\bm{x}_i^k,\bm{x}_j^k,{y_{ij}^k}} ) |i \neq j,k = 1,2, \ldots, K} \}$ as the SPS training set, where $k$ indicates the $k$-th series. The series photo selection problem is cast as a pair-wise ranking task. Let ${M_1}, {M_2}, \dots, {M_K}$ present the numbers of photos in $K$ series, then the size of $\mathcal{D}$ is $n=C^2_{M_1}+C^2_{M_1}+ \ldots+ C^2_{M_K}$. $\bm{x}_i^k$ and $\bm{x}_j^k$ are two different images in the $k$-th series, and ${y_{ij}^k}$ is a binary label indicating the winner, i.e., ${y_{ij}^k}=1$ when $\bm{x}_i^k$ is better than $\bm{x}_j^k$; otherwise, ${y_{ij}^k}=0$. In the following, we shall omit the subscript $k$ for notational simplicity.

% In our method, the final ranking is obtained by comparing the aesthetic quality of the images in pairs. For example, there are 3 images $I_1$, $I_2$, $I_3$ in a series. Through pairwise learning, we get $I_1>I_2$, $I_2>I_3$, then we can get the aesthetic quality ranking of all images in this series, that is, $I_1>I_2>I_3$.
We refer to the idea of \cite{Finkelstein16} quoting Bradley-Terry method \cite{1952Rank} to obtain the global ranking, and then predict the winning and losing probability when comparing two photos.
% We refer to the idea of Chang et al. \cite{Finkelstein16} quoting Bradley Terry model \cite{1952Rank} to obtain the global ranking, and then predict the winning and losing probability when comparing two photos.
\subsection{SPS-MVG}
Given image $\bm{x}$, we first established the feature extraction unit (FEU) to extract mutiple views. After we input the image $\bm{x}$ into this unit, the view matrix $V=[v^1,v^2,\ldots,v^l]\in R^{l \times C} $ is generated, where $l$ and $C$ denote the number of feature views and feature dimension. These models include, but are not limited to: manual feature extractors, deep learning extractors, general feature extractors, etc. To model relations between views, we construct an aesthetics-related graph based on the node representations. By performing the graph convolution, feature maps can be refined by message passing among nodes.

\subsubsection{Multi-view Graph Learning}

\textbf{Graph Building.} Considering $l$ nodes associated with the node representations $V$, we first construct an undirected fully connected graph $G=(V,E,A)$ in the coordinate space. Here, $G$ is constructed by its nodes $V$, the set of edges connecting nodes $E$ and adjacent matrix $A$ describing the edge weights. 
Since the features acquired by deep learning are far more effective in aesthetic assessment than hand-designed features and other general features, we regard the deep feature views as the central node of the graph. Then we calculate correlations between views to obtain the weight of the edges.
Specifically, we measure the content relations between views using the Cosine Similarity. 
After computing the pairwise similarity matrix, we obtain the affinity matrix $A \in\mathbb {R}^{l \times l}$ and adopt the softmax function for normalization on each row of the matrix by:
\begin{equation}
A_{pq} = \frac{exp(sim(v^p,v^q))}{\sum_{m=1}^l exp(sim(v^p,v^q))} ,
\end{equation}
where $exp(\cdot)$ is the exponential function. We set a threshold $\lambda$ to determine which edges are retained to exclude the negative effects caused by excessive feature differences.

\textbf{Graph Learning.} Suppose we encode the interactive information using a two-layer graph convolutional network,the latent graph interactive embeddings $F_G \in R^{n \times d}$ nodes can be constructed as follows:
\begin{equation}
H = \sigma(\widetilde{D}^{-\frac{{1}}{{2}}} \widetilde{A} \widetilde{D}^{-\frac{{1}}{{2}}}V W_{G_1}^T),
\end{equation}
\begin{equation}
F_G = \sigma(\widetilde{D}^{-\frac{{1}}{{2}}} \widetilde{A} \widetilde{D}^{-\frac{{1}}{{2}}}H W_{G_2}^T),
\end{equation}
Among them, $W_{G_1}^T$ and $W_{G_2}^T$ are layer-specific trainable weights of the control node encoder. $V$ is a matrix of node feature vectors $v_i$. $\sigma$ is the ReLU activation functions, $\widetilde{A}=A+I$ is the adjacency matrix of the undirected graph G, and $I$ is the identity matrix. $\widetilde{D}$ is diagonal matrix is defined as $\widetilde{D}=diag(d_1,d_2,...,d_n)$, $d_i=\sum_i{\widetilde{A}_{ij}}$.

\subsubsection{Self-attention module}

In order to obtain the correlation between the original features and highlight the important common features, the view features need to be fused for self-attention operation. After processing, we transform $V=[v^1,v^2,\ldots,v^l]\in R^{l \times C} $ into feature $F\in R^{l \times C}$.

The self-attention module can establish a transformation process to map input features $F \in\mathbb {R}^{M \times C}$ to a feature maps $F_S \in\mathbb {R}^{M \times C}$.
Given the input feature $F$, our proposed module first uses global average pooling and maximum pooling to squeeze the feature map along the spatial axis in parallel, then sum the features element by element. Finally, we use the \textit{Sigmoid} operation to get the channel attention matrix(vector).
  \begin{equation}
 \bm{a}=\delta(\bm{W_2} \sigma(\bm{W_1} \bm{(F_{Avg}+F_{Max})})),
 \end{equation}
where $\bm{a}\in\mathbb {R}^{1\times C}$ and $\delta(\cdot)$ denotes a \textit{Sigmoid} non-linearity function. 
Finally, the input feature $F$ is integrated with the channel attention $\bm{a}$ to get the channel-refined feature $F_S$.
\begin{equation}
\bm{F_S}= \bm{a} \circ \bm{F},
\end{equation}
where $\circ$ denotes element-wise multiplication. In this way, $\bm{F_S}$ can be regarded as an interactive feature selector, which is used to reinforce the common information between features for the next aesthetic quality assessment.

\subsection{Siamese architecture and Loss}
The siamese architecture is shown in Figure 3. Given a pair of images $(\bm{x}_i, \bm{x}_j)$ of the same series, the SPS-MVG network can obtain the feature representations $F_G^i + F_S^i$ of image $\bm{x}_i$ and $F_G^j + F_S^j$ of image $\bm{x}_j$ respectively. After connecting the features and passing through three fully connected layers, we can obtain the comparison result $f(\bm{x}_i, \bm{x}_j)$ of the image pair. $f(\bm{x}_i, \bm{x}_j)$ is a mapping function that predicts the probability of the preference of $\bm{x}_i$ over $\bm{x}_j$.
The desired mapping function can be obtained by minimizing the cross-entropy loss as follows:
\begin{equation}
L = \frac{1}{n}[\sum\limits_{k = 1}^n { - {y^k}\log f(\bm{x}_i^k,\bm{x}_j^k) - (1 - {y^k})} \log (1 - f(\bm{x}_i^k,\bm{x}_j^k))].
\end{equation}

\section{Experiments}
\subsection{Experimental setup}
\textbf{Datasets.} The Phototriage dataset \cite{Finkelstein16} is the first large-scale public dataset of unedited personal photos in the field of image aesthetics quality assessment research. The data set contains 15,545 photos organized in 5,953 series. In each series, there are approximately 2 to 8 photos, and the paired preference of the two photos is manually marked as ground truth. We shuffle and re-divide the training and validation sets of the dataset in units of series. After re-dividing, we randomly selected 10,023 photos for training, 2,499 photos for verification, and the remaining 2555 photos for testing.

\textbf{Implementation details.} Stochastic gradient decent (SGD) is used to optimize the network, and the weight decay is $10 ^{-5}$. On the Phototriage database, we used the $5 \times 10^{-2}$ learning rate and reduced the settings during the learning process. %Imagenet \cite{krizhevsky2012imagenet} is used to initialize the weights of the base deep network.

\textbf{Architecture.} We use the image feature extracted by VGG-16\footnote{We have also tried other deep learning methods such as ResNet and AlexNet, and VGG-16 yields the overall best performance.} \cite{Finkelstein16, SimonyanZ14a} as the center node, and other views are generated by other methods, e.g., SIFT, color histogram. The acquired features are only deep features for back propagation, and then a two-layer graph convolutional network \cite{KipfW17} and a self-attention module are used in parallel to process multi-view features, and finally three fully connected layers are processed and the classification results are obtained. For convenience of presentation, $\mathcal{V}$, $\mathcal{A}$ and $\mathcal{R}$ represents the VGG-16, AlexNet and ResNet-50, respectively. Similarly, $\mathcal{C}$ indicates the color, $\mathcal{H}$ means that whether hsv is involved, $\mathcal{S}$ uses the sift method in the ablation experiment. 

\textbf{Baselines.} We compare our algorithm with the following comparable baselines, including the Siamese network structure of many classic convolutional neural networks (such as VGG-16 \cite{Finkelstein16, SimonyanZ14a}, ResNet-50 \cite{HeZRS16}, AlexNet \cite{KrizhevskySH12}). And some methods of pairwise comparison (PAnet \cite{HuangCZSYY20} and PAC-Net \cite{KoLK18}).

\textbf{Evaluation metrics.} We use accuracy and F1 score to evaluate the test effects of each model on image pairs. 

\begin{table}[t]
\centering
\caption{Classification accuracy and F1 results on photo pairs.}
\scalebox{1.0}{
\begin{tabular}{c|c|c}
\hline
Method   & Accuracy(\%) & F1 score(\%) \\ \hline \hline
VGG-16 \cite{Finkelstein16, SimonyanZ14a}    & 72.322  & 72.153 \\ 
ResNet-50 \cite{HeZRS16}  & 69.037 & 68.668 \\ 
PAC-Net \cite{KoLK18}  & 71.623      & 71.062 \\ 
AlexNet \cite{KrizhevskySH12}  & 68.750      & 67.609 \\ 
PANet \cite{HuangCZSYY20}    & 71.362      & 71.353 \\ \hline
SPS-MVG      & \textbf{74.712}      & \textbf{74.193} \\ \hline
\end{tabular}
}
\end{table}

\begin{table}[t]
\centering
\caption{Results of the ablation study.}
\begin{tabular}{c|c|c|c}
\hline
Pooling                          & Views & Accuracy        & F1 Score        \\ \hline \hline
\multirow{3}{*}{null}    & $\mathcal{V}$ + $\mathcal{C}$ + $\mathcal{H}$     & 72.413          & 72.173          \\
                                 & $\mathcal{V}$ + $\mathcal{C}$ + $\mathcal{S}$     & 72.419          & 72.073          \\
                                 & $\mathcal{V}$ + $\mathcal{H}$ + $\mathcal{S}$     & 72.347          & 72.66           \\ \hline
\multirow{3}{*}{max}     & $\mathcal{V}$ + $\mathcal{C}$ + $\mathcal{H}$     & 73.922          & 73.367          \\
                                 & $\mathcal{V}$ + $\mathcal{C}$ + $\mathcal{S}$     & 73.204          & 72.793          \\
                                 & $\mathcal{V}$ + $\mathcal{H}$ + $\mathcal{S}$     & 73.131          & 72.7            \\ \hline
\multirow{3}{*}{avg}     & $\mathcal{V}$ + $\mathcal{C}$ + $\mathcal{H}$     & 72.988          & 72.186          \\
                                 & $\mathcal{V}$ + $\mathcal{C}$ + $\mathcal{S}$     & 71.91           & 71.186          \\
                                 & $\mathcal{V}$ + $\mathcal{H}$ + $\mathcal{S}$     & 71.623          & 71.272          \\ \hline
\multirow{3}{*}{max+avg} & $\mathcal{V}$ + $\mathcal{C}$ + $\mathcal{H}$     & 74.425          & 73.499          \\
                                 & $\mathcal{V}$ + $\mathcal{C}$ + $\mathcal{S}$     & \textbf{74.712} & \textbf{74.193} \\
                                 & $\mathcal{V}$ + $\mathcal{H}$ + $\mathcal{S}$     & 74.275          & 72.916          \\ \hline
\end{tabular}
\end{table}

\begin{figure*}[t]
\centering
\includegraphics[width=1\textwidth]{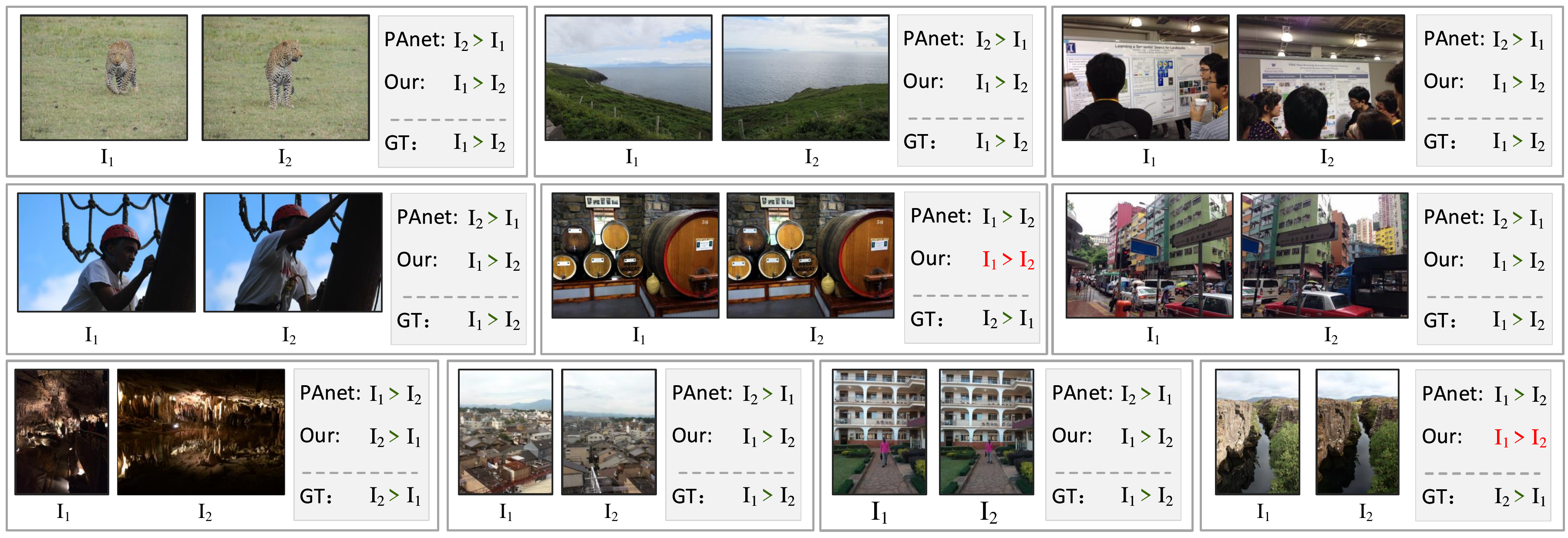}
\caption{Case study. We show some typical results here, and GT presents the ground-truth.}
\end{figure*}
% \caption{Several experimental results of the dataset. In comparison, most image pairs can be successfully distinguished, and a few are more difficult to distinguish.GT presents the ground-truth.}

\subsection{Experimental result}

Table 1 shows the comparison between and available advanced baselines. We evaluate the performance via the classification accuracy of the photo pairs. It is apparent that our method performs better than other approaches, which proves the effectiveness of our method for the problem of SPS. In Figure 4, we present a part of the results provided by SPS-MVG and PAnet. It is easy to observe that our comparison results for the same series of image pairs are improved compared to the previous method, but there are still some similar images that are difficult to distinguish successfully. The advantage of our method is that it takes into account the use of multi-view enhancement strategy. Exploiting the graph structure allows us to obtain the rational relationship between views, and the application of adaptive weighted attention strengthens the acquisition of the mutual information.

\begin{figure}[t]
\centering
\subfigure[]
{\includegraphics[width=0.23 \textwidth]{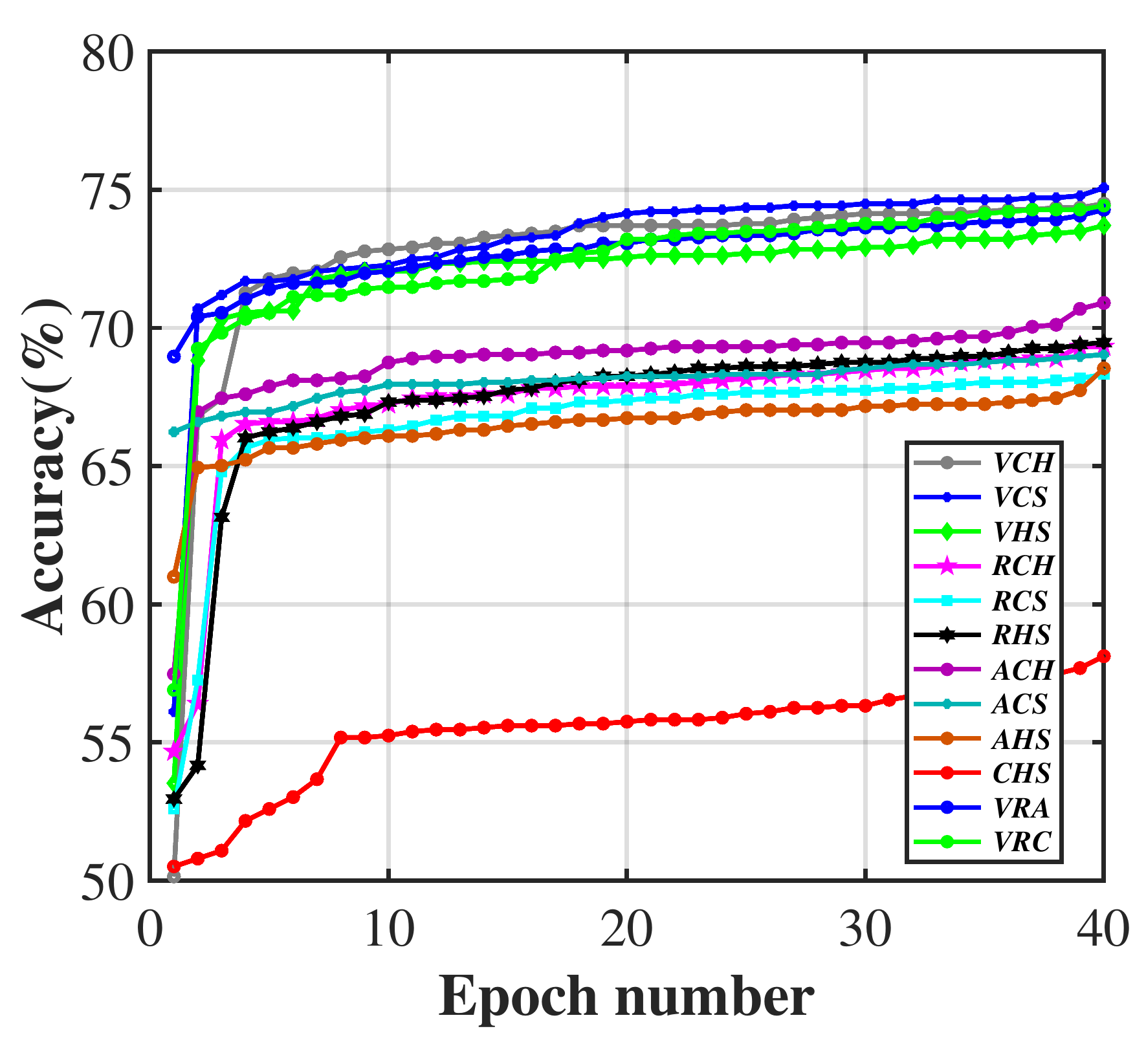}}
\subfigure[]
{\includegraphics[width=0.23 \textwidth]{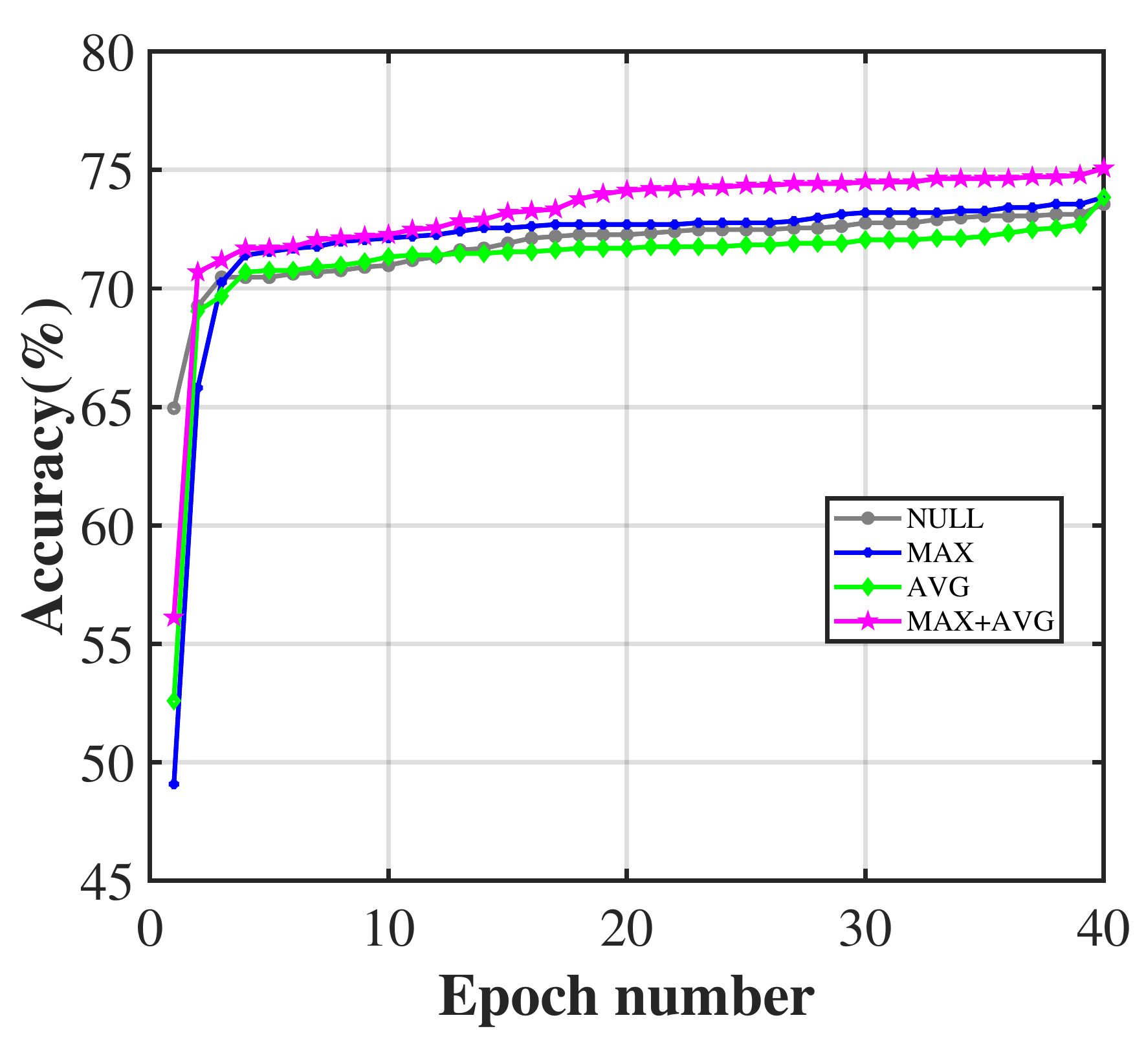}} 
\caption{Ablation study result curves over the validation set during training. (a) Our approach versus variant methods with different feature combinations. (b) Our approach versus variant methods with different pooling strategies.}
\end{figure}

\subsection{Ablation Study}
To analyses the contribution of each component of the final method SPS-MVG, all the ablation study results are shown in Tables 2 and Figure 5. 
% we analyze the ablation study here. 

\textbf{Multi-view Feature Combination.} In Figure 5(a), we implemented several variants of our method, which used features from different extraction methods for different combinations, and reported their accuracy.
% performance results with

On the one hand, Compared with the results in Table 1 and Figure 5(a), the experimental results of the method found that the multi-view effect of feature combination is better than the ordinary deep method, indicating that there is complementarity among the views, and comprehensive analysis can obtain a more comprehensive target object representation or description.

On the other hand, we take the combination of three features as an example to verify the effectiveness of the SPS-MVG method. We experimented with three combinations of shallow features, a combination of full deep features, a combination of a deep feature and two shallow features, a combination of two deep features and a shallow feature. Similarly, experiments with two and four features are also feasible. Among the Figure 5(a), the shallow training method fails to converge, indicating that the differences between the features are too large to obtain effective mutual information. The combination of full deep features is not as good as the effect of mixing features, indicating that there is complementary information between deep features and suitable shallow methods, and the combination of deep and shallow features can obtain the most comprehensive representation. The VGG network has the best effect among all the trained deep models, which is better than other ResNet and AlexNet frameworks, indicating that the VGG network is more suitable for solving the aesthetic quality assessment task of paired training. Only some features extracted by aesthetic methods are selected as an example. The combination of other features can achieve better results.

\textbf{Pooling Strategy Selection.} In our method, we compared whether to use the self-attention module, and the different pooling strategies equipped with the self-attention module, namely, the average pool strategy, the maximum pool strategy and their combined strategy. In this paper, only the features of the best VGG combination are used for testing. So we compare the above strategies in Figure 5(b). From Figure 5(b), we can see that the combination of maximum pooling and average pooling has the best effect. This confirms our view that the average pool and the maximum pool are complementary to each other. In order to improve accuracy, it is recommended to combine them.

\textbf{Model Stability Analysis.} Figure 5 displays the accuracy curves over the validation set of our approach and the variant methods during training in the above ablation studies. As can be observed, our approach leads to a monotonic improvement in the performance of the learning process, and it consistently outperforms the variant methods in all cases. The results further verify the stability of our approach. 

\section{Conclusion}
In this paper, we propose an effective series photo selection method, named SPS-MVG, based on the multi-view graph learning. The combination of deep and manual feature views can effectively reflect subtle aesthetic changes. To enhance multi-view correlations, we first propose a multi-view graph learning network to construct the spatial relationships between feature views. Through the aggregation of multiple views and self-attention modules with adaptive weights, we verified the importance of each view and obtained a more comprehensive image representation. At last, a siamese network is proposed to select the best one from a series of nearly identical photos. Extensive evaluations show that our proposed model achieves the best performance on the benchmark visual aesthetics dataset.

% \section{Citations and References}

% List and number all bibliographical references at the end of the paper. The references can be numbered in alphabetic order or in order of appearance in the document. When referring to them in the text, type the corresponding reference number in square brackets as shown at the end of this sentence~\cite{Morgan2005}. All citations must be adhered to IEEE format and style. Examples such as~\cite{Morgan2005},~\cite{cooley65} and~\cite{haykin02} are given in Section 12.

% \bibliographystyle{IEEEbib}
% \bibliography{icme2021template}

\end{document}